\documentclass[sigconf]{acmart}
\AtBeginDocument{%
  }

\usepackage{multirow}
\usepackage{algorithmic}
\usepackage{algorithm}
\usepackage{colortbl}

\setcopyright{acmlicensed}
\copyrightyear{2025}
\acmYear{2025}
\setcopyright{acmlicensed}\acmConference[MM '25]{Proceedings of the 33rd ACM International Conference on Multimedia}{October 27--31, 2025}{Dublin, Ireland}
\acmBooktitle{Proceedings of the 33rd ACM International Conference on Multimedia (MM '25), October 27--31, 2025, Dublin, Ireland}
\acmDOI{10.1145/3746027.3755792}
\acmISBN{979-8-4007-2035-2/2025/10}

\begin{document}

\title{VISA: Group-wise Visual Token Selection and Aggregation via Graph Summarization for Efficient MLLMs Inference}

\author{Pengfei Jiang}
\authornote{This work was done during an internship at Tencent Youtu Lab.}
\affiliation{%
  \institution{Key Laboratory of Multimedia Trusted Perception and Efficient Computing, Ministry of Education of China, Xiamen University}
  \city{Xiamen}
  \country{China}
}

\author{Hanjun Li}
\affiliation{%
  \institution{Tencent Youtu Lab}
  \city{Shenzhen}
  \country{China}
}

\author{Linglan Zhao}
\affiliation{%
  \institution{Tencent Youtu Lab}
  \city{Shanghai}
  \country{China}
}

\author{Fei Chao}
\authornote{Corresponding authors.}
\affiliation{%
  \institution{Key Laboratory of Multimedia Trusted Perception and Efficient Computing, Ministry of Education of China, Xiamen University}
  \city{Xiamen}
  \country{China}
}

\author{Ke Yan}
\authornotemark[2]
\affiliation{%
  \institution{Tencent Youtu Lab}
  \city{Shanghai}
  \country{China}
}

\author{Shouhong Ding}
\affiliation{%
  \institution{Tencent Youtu Lab}
  \city{Shanghai}
  \country{China}
}

\author{Rongrong Ji}
\affiliation{%
  \institution{Key Laboratory of Multimedia Trusted Perception and Efficient Computing, Ministry of Education of China, Xiamen University}
  \city{Xiamen}
  \country{China}
}

\renewcommand{\shortauthors}{Pengfei Jiang et al.}

\begin{abstract}
In this study, we introduce a novel method called group-wise \textbf{VI}sual token \textbf{S}election and \textbf{A}ggregation (VISA) to address the issue of inefficient inference stemming from excessive visual tokens in multimoal large language models (MLLMs).
Compared with previous token pruning approaches, our method can preserve more visual information while compressing visual tokens.
We first propose a graph-based visual token aggregation (VTA) module. VTA treats each visual token as a node, forming a graph based on semantic similarity among visual tokens. It then aggregates information from removed tokens into kept tokens based on this graph, producing a more compact visual token representation.
Additionally, we introduce a group-wise token selection strategy (GTS) to divide visual tokens into kept and removed ones, guided by text tokens from the final layers of each group. 
This strategy progressively aggregates visual information, enhancing the stability of the visual information extraction process.
We conduct comprehensive experiments on LLaVA-1.5, LLaVA-NeXT, and Video-LLaVA across various benchmarks to validate the efficacy of VISA. Our method consistently outperforms previous methods, achieving a superior trade-off between model performance and inference speed.
The code is available at \url{https://github.com/mobiushy/VISA}.
\end{abstract}





\maketitle

\begin{figure*}[!t]
  \centering
  \includegraphics[width=\textwidth]{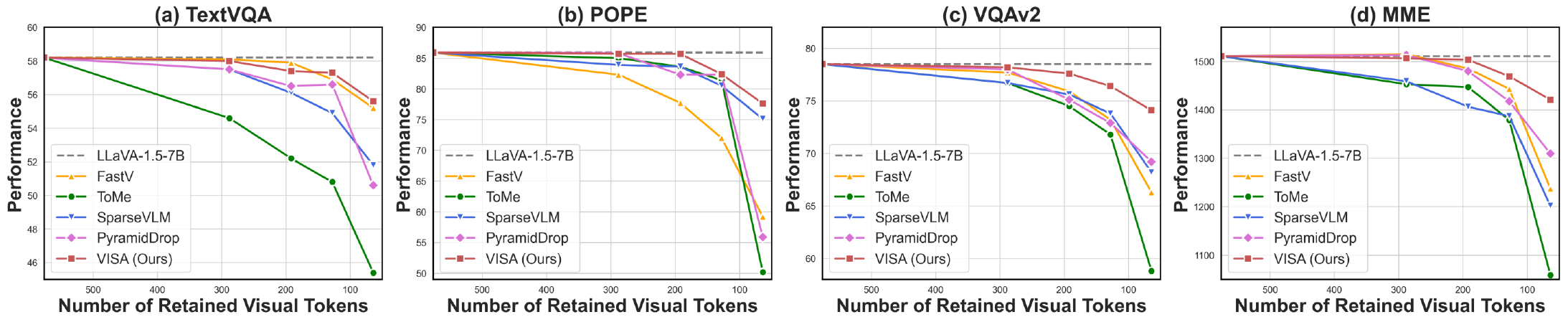}
  \caption{Performance comparison on 4 widely-used multimodal benchmarks. Our method can prevent excessive visual information loss by aggregating visual tokens, thereby achieving superior trade-off between model performance and speed.}
  \label{fig1}
\end{figure*}

\section{Introduction}

Recently, the significant success of Large Language Models (LLMs) \cite{ref1, ref2, ref3, ref4, ref5} in natural language processing has spurred the growth of the multimodal field. To equip LLMs with visual processing capabilities, \citet{ref6} first extract image features through a visual encoder~\cite{ref21}. They then devised an MLP projector to align image features with the semantic space of the LLM, treating them as visual tokens. These visual tokens are then concatenated with text tokens and fed into the LLM, creating a multimodal large language model.
While MLLMs~\cite{ref6, ref7, ref8, ref9, ref10} have excelled in tasks like image understanding and visual question answering, excessive visual tokens causes significant computational overhead, hampering the efficiency of inference. For instance, in LLaVA-1.5~\cite{ref6}, $336\times336$ images are transformed into 576 tokens, accounting for roughly 64\% of the input tokens. Some studies~\cite{ref37, ref38} have found that the model performance can be further enhanced by increasing the resolution of the input image. Other studies~\cite{ref16, ref43, ref44} have proposed using video as the visual modality input to support video understanding tasks. However, with high-resolution images or video input, the visual token count increases further. LLaVA-NeXT~\cite{ref8} can handle up to 2880 visual tokens, while Video-LLaVA~\cite{ref16} selects 8 frames from each video input and encodes them into 2048 visual token inputs. To handle excessive visual token input, \cite{ref9, ref17, ref18} extract more compact image representations by modifying the image encoder or projector. However, these new model structures require retraining the entire model to achieve efficient inference, which also increases the computational overhead.

Some training-free token compression techniques~\cite{ref11, ref12, ref19, ref20} have been applied to MLLMs. These approaches do not need extra fine-tuning and can be applied plug-and-play to pre-trained MLLMs while maintaining specific model performance.
FastV~\cite{ref11} initially identified the inefficient phenomenon of visual attention in MLLMs and utilized the attention score as an indicator of the importance of the visual token. Subsequently, visual tokens with low attention scores will be discarded directly to reduce redundancy. However, directly discarding visual tokens will inevitably result in the loss of visual information, impacting the model's performance. 
SparseVLM~\cite{ref19} employs both token pruning and merging techniques. Initially, it prunes insignificant visual tokens following the question prompt's guidance. Subsequently, it introduces a token recycling mechanism to prevent excessive loss of visual information. However, this approach still fails to preserve sufficient visual information, leading to a significant deterioration in model performance under high token compression rates, as illustrated in Figure\,\ref{fig1}. In the realm of ViT's token compression, ToMe~\cite{ref12} proposed progressively merging tokens instead of pruning them to prevent information loss. In this way, it still achieved competitive performance after pruning a large number of tokens on image classification tasks. However, Zhang~\cite{ref19} first applied ToMe in MLLM inference for comparison and observed that it failed to maintain performance on multimodal understanding tasks. Actually, ToMe~\cite{ref12} adopted a layer-wise token merging algorithm by averaging feature to gradually merge similar tokens together, thereby reducing the number of tokens and speeding up inference. However, merging tokens by averaging features will result in the merged token being a coarse-grained representation of these similar tokens, leading to the loss of detailed image information. While this coarse-grained representation works well for image classification tasks, it is unsuitable for tasks like visual question answering that demand detailed image information, leading to a significant drop in performance. In our study, as depicted in Figure\,\ref{fig1}, ToMe~\cite{ref12} does not perform well in tasks that require detailed image comprehension, such as TextVQA.

We argue that the progressive pruning and merging strategy is promising. To avoid the aforementioned drawbacks of ToMe, it requires an elaborate information aggregation mechanism for those pruned tokens instead of simple averaging features.
In this paper, we introduce a novel method called group-wise \textbf{VIsual token Selection and Aggregation (VISA)} to overcome the limitations of previous methods and achieve superior performance-speed balance. 


Drawing inspiration from graph summarization techniques~\cite{ref13, ref14, ref15}, which aggregate node information into super nodes through a graph network for a more compact graph representation, we first propose a graph-based visual token aggregation (VTA) module. The VTA treats each visual token as a node and construct a graph based on the semantic similarity of visual tokens. It then aggregates the information from less critical visual tokens into the important ones (super nodes) based on this graph network, and discards these less critical visual tokens. Therefore, visual information is summarized into super nodes, offering a more compact representation of image features while preserving maximum visual information. Meanwhile, we introduce a group-wise token selection strategy, which aims to provide reliable division of visual tokens for VTA and ensure more stable extraction of visual information. Specifically, we partition the layers of LLM into groups and apply VTA solely at the end of each group. Different from FastV~\cite{ref11}, we consider multi-layer attention scores to distinguish critical visual tokens and less important ones. Instead of applying VTA at single layer like FastV or every layer like ToMe, VTA only aggregates and prunes tokens at the last layer of each group in a progressive manner throughout the LLM. In this way, the kept tokens can steadily assimilate the information from removed ones before pruning in next group. 

In summary, the contributions of this paper include:

\textbullet ~ We propose a graph-based visual token aggregation algorithm that propagates the information from less critical visual tokens to important ones through graph summarization to obtain compact visual token representation while retaining visual information.

\textbullet ~ We propose a group-wise token selection strategy to divide visual tokens into kept and removed ones guided by text tokens from the final layers of each group. This strategy ensures stability and accuracy in visual information extraction and token selection.

\textbullet ~ Extensive experiments demonstrate that our method consistently outperforms previous approaches. Notably, our method can speed up inference by 2.08 times on LLaVA-1.5-13B while maintaining 98.14\% of the model's performance.

\begin{figure*}[!t]
  \centering
  \includegraphics[width=\textwidth]{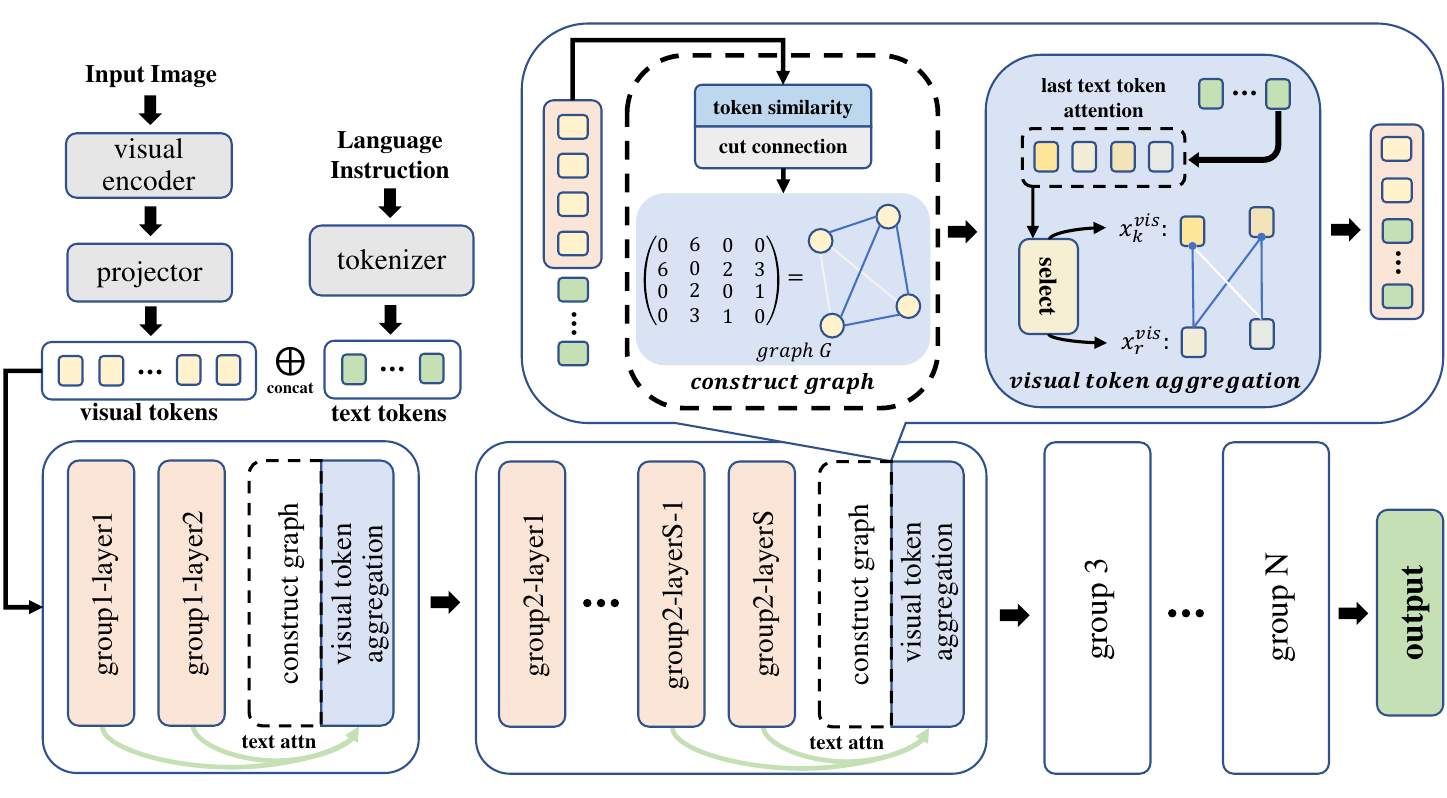}
  \caption{The framework of VISA. We divide LLM layers into multiple groups, and visual token aggregation (VTA) is apply to the end of each group. VTA first construct graph based on the semantic similarity among visual tokens, then utilize this graph to aggragate visual tokens through graph summarization. 
  Additionally, a group-wise token selection strategy (GTS) is proposed to select $x^{vis}_{k}$ from $x^{vis}$ based on text attention guidance from the final layers of each group.
  }
  \vspace{-0.2em}
  \label{fig2}
\end{figure*}

\section{Related Work}

\subsection{Multimodal Large Language Models}
To empower Large Language Models (LLMs) with visual understanding abilities, recent studies first employ a visual encoder to extract visual features and then an alignment module is used to align visual features with textual features. For instance, BLIP-2~\cite{ref39} utilizes the FlanT5 model for multimodal understanding and employs Q-Former as a projector to bridge the modality gap. 
Qwen-VL~\cite{ref40} also employs a Q-Former projector to align visual and textual modalities.
However, using Q-Former as the alignment module will lead to substantial training overhead.
LLaVA~\cite{ref41} employs a simple linear projector to align visual and textual modalities. This linear projector and LLM are fine-tuned on a high-quality visual instruction tuning dataset. In order to further make up for the visual shortcomings of MLLMs and alleviate the hallucination problem, some studies have enhanced the model performance by increasing the resolution of the input image. 
For instance, LLaVA-1.5~\cite{ref6} further enhances the vision encoder to accommodate higher resolution image input and substitutes the linear projector with a multi-layer perceptron. This results in more visual tokens being fed into the language model. On the other hand, LLaVA-NeXT~\cite{ref8} doubles the image resolution, further increasing the number of visual tokens and the model's inference overhead. Hence, simplifying visual token representation to speed up model inference has emerged as a pressing topic.

\subsection{Visual Token Reduction for MLLMs}
In order to shorten the length of visual tokens, token pruning methods directly discard unimportant tokens. For example, FastV~\cite{ref11} discards less important visual tokens based on their attention scores, while VTW~\cite{ref20} directly removes all visual tokens at a certain layer of LLM. Token merging~\cite{ref12} is an alternative method to shorten tokens, which can avoid the information loss caused by directly discarding tokens to a certain extent. Meanwhile, there are some methods that utilize both token pruning and token merging techniques. LLaVA-PurMerge~\cite{ref42} identifies crucial visual tokens with high attention scores and merges them through similar key clustering, enhancing model inference efficiency and achieving competitive performance. 
FasterVLM~\cite{ref47} prunes the visual tokens before entering the projector. It re-ranks image tokens using cls attention from the visual encoder and prune the last $R\%$.
SparseVLM~\cite{ref36} first selects text raters based on the correlation between text tokens and input image. 
It then uses their attention scores on visual tokens to identify important ones and introduces a token recycling mechanism to reduce information loss caused by discarding tokens. 
PyramidDrop~\cite{ref46} divide the LLM forward pass into multiple stages, and drop part of the image tokens at the end of each stage with a pre-defined ratio based on the attention score of the image tokens.
However, these methods still struggle to preserve sufficient visual information under high visual token compression rates, resulting in a significant degradation in model performance.

\section{Methodology}

\subsection{Preliminary}

Existing MLLMs typically comprise three components: a visual encoder, projector, and LLM decoder. Since our method is utilized in LLM decoder, we begin by presenting LLM and the symbols used.
The LLM decoder utilizes a transformer-based architecture~\cite{ref22}, consisting of multiple transformer blocks. Each transformer block consists primarily of causal self-attention and a feed-forward network. The input visual and text signals are converted into visual and text tokens, respectively, which are concatenated together and sent to the LLM decoder to generate a response.

The input token sequence $X=[x^{sys};\,x^{vis};\,x^{txt};\,x^{out}] \in \mathbb{R}^{l \times d}$ is first converted into query $Q$, Key $K$, and value $V$ by three weight matrices $W_Q, W_K, W_V \in \mathbb{R}^{d \times d}$ in each transformer block. Here, $l$ is the total token length and $d$ is the feature dimension size. Then, the causal self attention is calculated as follows:
\begin{equation}
\label{eq1}
    A = Softmax(\frac{QK^{T} + M}{\sqrt{d}}), ~~~ Y = AV
\end{equation}
where $M \in \mathbb{R}^{l \times l}$ is a lower triangular causal mask ensuring that each token attends only to itself and previous tokens.
Then the hidden states $Y$ are fed into the feed-forward network to get the output $O = FFN(Y)$ of the transformer block. After passing through all the transformer blocks, the currently predicted token is obtained. This token is then concatenated with the previous tokens and sent to the LLM decoder again to predict the next token:
\begin{equation}
\label{eq2}
    p(y) = \prod^{l_y}_{i=1}p(y_i~|~X;y_{i-1})
\end{equation}
Here, $y$ represents the model's predicted answer, while $l_y$ denotes the token length of answer $y$. The LLM decoder predicts the answer $y$ in an autoregressive manner.

\begin{algorithm}[tb]
   \caption{Visual Token Selection and Aggregation}
   \label{alg1}
    \begin{algorithmic}[1]
       \STATE {\bfseries Input:} visual tokens $x^{vis}$, text tokens $x^{txt}$, kept token rate $p$, the number of groups $N$
       \FOR{$g=1$ {\bfseries to} $N$}
           \STATE $G_{i,j} = \cos(x^{vis}_{i},x^{vis}_{j})$ \COMMENT{Construct graph $G$}
           
           \STATE $G_{i,j} = 0, ~~~ if~i=j~or~G_{i,j} < 0$
           
           \STATE Symmetrically normalize $G$ according to Eq.~(\ref{eq4})
           
           \STATE $I = \frac{1}{L} \sum_{i} \sum_{j} A^{t2v}_{i,j}$
           \COMMENT{Calculate indictor $I$}
           
           \STATE $x^{vis}_{k} = topk(x^{vis}, I, p)$
           \STATE $x^{vis}_{r} = x^{vis} \setminus x^{vis}_{k}$
           \COMMENT{Split $x^{vis}$ according to $I$}
           
           \STATE Perform graph summarization according to Eq.~(\ref{eq5})
           
           \STATE $x^{vis} = x^{vis}_{k}$
           \COMMENT{Update visual tokens}
           
           \STATE $[x^{vis};x^{txt}] = LLM_{g}([x^{vis};x^{txt}])$
       \ENDFOR
       \STATE {\bfseries Output:} visual tokens $x^{vis}$, text tokens $x^{txt}$
       
    \end{algorithmic}
\end{algorithm}

\subsection{Visual Token Aggregation}
\label{sec3.2}

As illustrated in Figure\,\ref{fig2}, we divide the layers of LLM into groups (see in Sec\,\ref{sec3.3}) and apply VTA at the end of each group. In this section, we will take one of the Visual Token Aggregation modules as an example to explain the specific details.
VTA considers each visual token as a node and constructs a graph based on their semantic similarities. The information of the removed nodes is then aggregated into the kept nodes according to this graph network, resulting in a more compact visual token representation.

Specifically, given a visual token sequence $x^{vis} \in \mathbb{R}^{n \times d}$, where $n$ is the length of visual tokens. We construct a graph based on the cosine similarity between visual tokens to represent the weights of the edges between nodes. The adjacency matrix $G$ of this graph is defined as:
\begin{equation}
\label{eq3}
    {G}_{i,j} = \left\{ 
        \begin{array}{lc}
            \max(\cos(x^{vis}_{i},x^{vis}_{j}),~0) & i \neq j, \\
            0 & otherwise \\
        \end{array}
    \right.
\end{equation}
In order to prevent propagating the node's own information to itself, we reset the edge weight between a node and itself to 0. Additionally, we also disable any edge with a cosine similarity below 0 to avoid aggregating information from nodes with distinct semantics.
Then, we symmetrically normalize the adjacency matrix $G$ as:
\begin{equation}
\label{eq4}
    \hat{G} = D^{-\frac{1}{2}}GD^{-\frac{1}{2}}, ~~~ D_{i,i} = \textstyle{\sum_j} G_{i,j}
\end{equation}
where $D$ is the diagonal degree matrix of $G$ and the value on the diagonal represents the degree of the $i$-th node.

After completing the construction of the graph, we split the visual tokens $x^{vis}$ into two sets: the kept visual token set $x^{vis}_{k} \in \mathbb{R}^{n_k \times d}$ and the removed visual token set $x^{vis}_{r} \in \mathbb{R}^{n_r \times d}$. Here, $n_k$ and $n_r$ represent the length of kept and removed visual tokens, respectively. The division manner will be introduced in detail in Sec\,\ref{sec3.3}. Subsequently, through graph summarization technology, we aggregate the information of the removed visual tokens into the kept visual tokens that are semantically connected with them:
\begin{equation}
\label{eq5}
    x^{vis}_{k} = x^{vis}_{k} + \alpha \cdot \hat{G}_{A} x^{vis}_{r}
\end{equation}
\begin{table*}[!t]
  \centering
  \setlength{\tabcolsep}{3.5pt}
  \renewcommand\arraystretch{1.25}
  \caption{Comparison with previous approaches on LLaVA-1.5-7B across 10 multimodal benchmarks. Our method consistently outperform competitors at all kept visual token number settings. Even with 88.9\% of visual tokens pruned, our method maintains 93.8\% performance.}
  \begin{tabular}{@{}lcccccccccccc@{}}
    \toprule
    Method & TFLOPs & MM-Vet & MMB & MMB-CN & POPE & SQA-IMG & TextVQA & VQAv2 & VizWiz & MME & GQA & Avg. \\
    \midrule
    LLaVA-1.5-7B & 9.43 & 31.1 & 64.3 & 58.3 & 85.9 & 66.8 & 58.2 & 78.5 & 50.0 & 1510.7 & 62.0 & 100\% \\
    \midrule
    \rowcolor{gray!20}\multicolumn{13}{c}{\emph{Retain 192 Tokens}} \\
    FastV & \multirow{5}{*}{4.59} & 29.8 & 64.2 & 57.6 & 77.7 & 67.7 & 57.9 & 75.9 & 50.8 & 1485.1 & 57.9 & 97.6\% \\
    ToMe &  & 28.8 & 63.3 & 57.6 & 83.6 & 68.3 & 52.2 & 74.5 & 50.8 & 1446.7 & 58.5 & 96.6\% \\
    SpareseVLM &  & 31.2 & 62.5 & 56.1 & 83.6 & 69.1 & 56.1 & 75.6 & 51.1 & 1406.2 & 57.6 & 97.5\% \\
    PyramidDrop &  & 31.2 & 63.3 & 56.8 & 82.3 & 68.8 & 56.5 & 75.1 & 51.1 & 1479.7 & 57.3 & 98.0\% \\
    VISA (Ours) &  & 31.7 & 64.5 & 58.7 & 85.7 & 67.9 & 57.4 & 77.6 & 49.7 & 1503.3 & 60.3 & \textbf{99.8\%} \\
    \midrule
    \rowcolor{gray!20}\multicolumn{13}{c}{\emph{Retain 128 Tokens}} \\
    FastV & \multirow{5}{*}{3.80} & 26.8 & 63.3 & 56.3 & 72.0 & 68.0 & 56.9 & 73.2 & 51.3 & 1442.9 & 55.8 & 94.6\% \\
    ToMe &  & 25.3 & 61.8 & 56.5 & 81.4 & 67.8 & 50.8 & 71.8 & 50.3 & 1378.8 & 57.8 & 93.4\% \\
    SpareseVLM &  & 30.1 & 60.0 & 53.9 & 80.5 & 67.1 & 54.9 & 73.8 & 52.2 & 1386.9 & 56.0 & 95.2\% \\
    PyramidDrop &  & 29.4 & 61.6 & 56.6 & 82.3 & 68.4 & 56.6 & 72.9 & 51.0 & 1417.3 & 57.1 & 96.4\% \\
    VISA (Ours) &  & 30.2 & 64.8 & 58.0 & 82.4 & 67.8 & 57.3 & 76.4 & 49.2 & 1469.0 & 58.2 & \textbf{98.0\%} \\
    \midrule
    \rowcolor{gray!20}\multicolumn{13}{c}{\emph{Retain 64 Tokens}} \\
    FastV & \multirow{5}{*}{3.01} & 27.3 & 59.5 & 52.1 & 59.2 & 68.1 & 55.2 & 66.3 & 51.8 & 1237.1 & 51.6 & 88.9\% \\
    ToMe &  & 19.9 & 48.9 & 37.9 & 50.2 & 68.8 & 45.4 & 58.8 & 49.1 & 1059.0 & 49.8 & 76.8\% \\
    SpareseVLM &  & 22.3 & 56.2 & 46.5 & 75.1 & 62.2 & 51.8 & 68.2 & 51.3 & 1201.4 & 52.7 & 86.2\% \\
    PyramidDrop &  & 25.9 & 58.8 & 50.5 & 55.9 & 69.0 & 50.6 & 69.2 & 50.7 & 1309.2 & 47.5 & 86.9\% \\
    VISA (Ours) &  & 25.0 & 62.1 & 57.3 & 77.6 & 67.9 & 55.6 & 74.1 & 47.9 & 1420.6 & 56.2 & \textbf{93.8\%} \\
  \bottomrule
  \end{tabular}
  \label{table1}
\end{table*}
where $\hat{G}_{A} \in \mathbb{R}^{n_k \times n_r}$ represents the subgraph of the symmetrically normalized adjacency matrix $\hat{G}$. In matrix $\hat{G}_{A}$, the row indices and column indices correspond to the kept tokens and removed tokens, respectively. In other words, each row in $\hat{G}_{A}$ represents the relevance of a retained token to all removed tokens, that is, the information of each removed token can be propagated to multiple different tokens, which is the difference from the token merging~\cite{ref12} method.
The hyperparameter $\alpha$ determines the magnitude of information propagation.

We implement visual token aggregation between each group of transformer layers in the LLM decoder. Each time token aggregation is performed, a new graph is constructed based on the current updated visual tokens. This dynamic graph construction process ensures that the updated visual token information can be utilized when aggregating visual tokens.

\noindent\textbf{Disccusion.} Previous studies~\cite{ref48, ref49, ref15} have utilized graph-based methods for token pruning. Our approach differs from these methods both in technical details and application scenarios. Specifically, Zero-TPrune~\cite{ref48} and G-Prune~\cite{ref49} employ an iterative graph propagation process to determine each token's importance score and then discard tokens with low scores. In contrast, our method directly propagates the information of the token itself to aggregate visual tokens. Unlike GTP-ViT~\cite{ref15}, which focuses on token compression in ViT, our method is tailored for MLLMs. Additionally, our graph construction and token aggregation occur between each group of layers rather than between self-attention and feed forward network in each transformer block as in \cite{ref15}. Moreover, we consider the text attention information across the entire group when dividing visual tokens. In contrast, GTP-ViT relies solely on image modality information to score each visual token. These designs facilitate our method in achieving superior model performance and in being more suitable for multimodal large language models.


\subsection{Group-wise Token Selection Strategy}
\label{sec3.3}

In this section, we will explain how to apply VTA to MLLMs, including when to aggregate token information and how to split $x^{vis}_{k}$ and $x^{vis}_{r}$. 
Different from FastV's one-time pruning strategy in the second layer of LLM and ToMe's layer-wise merging strategy, we suggest dividing the LLM layers into groups and conducting visual token aggregation at the end of each group.
This strategy ensures that each time the visual token representation is modified, the LLM has sufficient layers to extract visual information from the updated representation, enhancing the stability of the information extraction process.
The initial group comprises the first two layers to facilitate global information extraction across all tokens in these layers, which is evidenced in \cite{ref11}. Subsequently, each successive $S$ layer is treated as an individual group starting from the second layer. Here, $S$ is a hyperparameter that can be adjusted.

For the division manner, we utilize attention information from the entire group to enhance the effectiveness of partitioning $x^{vis}_{k}$ and $x^{vis}_{r}$. Specifically, we average the attention scores from the final $L$ layers of each group to guide the division of kept tokens and the removed tokens. Meanwhile, given that MLLMs determine the focal area of the image based on the question prompt, we utilize the attention score from the last text token to the visual token as an indicator of the visual token's importance, denoted as $I$:
\begin{equation}
\label{eq6}
    I = \frac{1}{L \cdot H} \sum_{i} \sum_{j} A^{t2v}_{i,j}
\end{equation}
where $A^{t2v}_{i,j} \in \mathbb{R}^{1 \times n}$ represents the attention score of the $j$-th head of the last text token on the visual tokens in the $i$-th layer. We average the attention scores of all heads $H$ to ensure that information from all subspaces is utilized. 
Then, we employ metric $I$ to divide $x^{vis}$ into kept visual tokens $x^{vis}_{k}$ and removed visual tokens $x^{vis}_{r}$:
\begin{equation}
\label{eq7}
    x^{vis}_{k} = topk(x^{vis}, I, p), ~~~ x^{vis}_{r} = x^{vis} \setminus x^{vis}_{k}
\end{equation}
Here, $topk(\cdot)$ selects the proportion $p$ of visual tokens with the highest importance indicator $I$. The removed visual tokens $x^{vis}_{r}$ is the difference between the total visual tokens $x^{vis}$ and the kept visual tokens $x^{vis}_{k}$.
By adjusting the quantity or proportion of kept visual tokens and the number of layers $S$ in each group, our method can be tailored to various FLOPs requirements.

The seamless integration of the group-wise token selection strategy and the visual token aggregation in Sec\,\ref{sec3.2} ensures that visual tokens in LLM are progressively transformed into more compact visual representations, striking a balance between model performance and inference speed. The overall algorithm of our method is illustrated in Algorithm~\ref{alg1}.

\subsection{Computational Complexity Analysis}
\label{sec3.4}
Here we analyze the computational complexity of our method and the computational overhead saved by our method. The extra computational cost from our method primarily occurs during graph construction and visual token aggregation. However, this computation is minimal compared to the transformer layers. Denoting the proportion of tokens retained in each VTA process as $p$, the additional overhead introduced by our method is $O((p \cdot n)^2)$. As visual tokens are progressively aggregated, this overhead diminishes. Meawhile, since we conduct VTA between each group of layers, the extra computation only happens $N-1$ times.

Next, we analyze the amount of computation saved by our method. Given an input token sequence $x \in \mathbb{R}^{n \times d}$, the computational complexity of a transformer layer is $O(n^2d+nd^2)$. To simplify the analysis, we approximate the computational complexity of a transformer layer as proportional to $n$, denoted as $O(c \cdot n)$ where $c=nd+d^2$. Therefore, the computational cost of our method is:
\begin{equation}
\label{eq8}
    \frac{(1-p^N)}{N(1-p)} \cdot L \cdot c \cdot n
\end{equation}
Where $N$ represents the number of groups and $L$ indicates the number of LLM layers. It can be observed that compared to the total computational load $L \cdot c \cdot n$ of the original LLM decoder, by controlling the hyperparameters $p$ and $N$, the computational overhead of our method is $(1-p^N)/N(1-p)$ of original.

\begin{table}[!t]
  \centering
  \setlength{\tabcolsep}{3pt}
  \renewcommand\arraystretch{1.2}
  \caption{Comparison with previous methods on LLaVA-1.5-13B, the average performance on 10 benchmarks is presented.}
  \begin{tabular}{@{}lcccccc@{}}
    \toprule
    Method & Tokens & Avg. & Tokens & Avg. & Tokens & Avg. \\
    \midrule
    FastV & \multirow{5}{*}{192} & 97.9\% & \multirow{5}{*}{128} & 96.0\% & \multirow{5}{*}{64} & 91.3\% \\
    ToMe &  & 97.1\% &  & 92.4\% &  & 86.2\% \\
    SparseVLM &  & 97.0\% &  & 95.3\% &  & 90.2\% \\
    PyramidDrop &  & 98.9\% &  & 97.3\% &  & 92.9\% \\
    VISA (Ours) &  & \textbf{99.2\%} &  & \textbf{98.1\%} &  & \textbf{94.3\%} \\
    
  \bottomrule
  \end{tabular}
  \label{table2}
\end{table}

\begin{table}[!t]
  \centering
  \setlength{\tabcolsep}{3pt}
  \renewcommand\arraystretch{1.2}
  \caption{Comparison with previous methods on LLaVA-NeXT-7B, the average performance on 10 benchmarks is presented.}
  \begin{tabular}{@{}lcccccc@{}}
    \toprule
    Method & Tokens & Avg. & Tokens & Avg. & Tokens & Avg. \\
    \midrule
    FastV & \multirow{5}{*}{720} & 98.0\% & \multirow{5}{*}{360} & 94.0\% & \multirow{5}{*}{180} & 85.1\% \\
    ToMe &  & 91.2\% &  & 87.8\% &  & 66.1\% \\
    SparseVLM &  & 94.3\% &  & 91.2\% &  & 77.7\% \\
    PyramidDrop &  & 97.1\% &  & 93.3\% &  & 88.8\% \\
    VISA (Ours) &  & \textbf{99.5\%} &  & \textbf{95.6\%} &  & \textbf{91.1\%} \\
    
  \bottomrule
  \end{tabular}
  \label{table3}
\end{table}

\section{Experiments}

In this section, we conduct comprehensive experiments to validate our method across various scenarios.
Meanwhile, efficiency analysis is performed to demonstrate the acceleration effect of our method. Finally, we further verify our method through visualization results and ablation study.

\subsection{Experimental Setup}

\textbf{Benchmarks and metrics.}
We evaluate our method and other competitors on 10 image-based multi-modal benchmarks, including TextVQA~\cite{ref23}, SQA-IMG~\cite{ref24}, VizWiz~\cite{ref25}, VQAv2~\cite{ref26}, and GQA~\cite{ref27}, as well as MM-Vet~\cite{ref28}, MMBench~\cite{ref29}, MMBench-CN~\cite{ref29}, POPE~\cite{ref30}, and MME~\cite{ref31}. 
For video understanding, we evaluate on TGIF-QA~\cite{ref33}, MSVD-QA~\cite{ref34}, MSRVTT-QA~\cite{ref34}, and ActivityNet-QA~\cite{ref35}. All experiments on these benchmarks follow the default settings and evaluation metrics.

\begin{table}[!t]
  \centering
  \setlength{\tabcolsep}{8pt}
  \renewcommand\arraystretch{1.2}
  \caption{Inference efficiency analysis on MMBench.}
  \begin{tabular}{@{}lccc@{}}
    \toprule
    Mehod & Tokens & Total time & Samples/sec \\
    \toprule
    LLaVA-1.5-7B & 576 & 19m45s & 3.69 \\
    \midrule
    \multirow{3}{*}{w/ fastv} & 192 & 14m04s & 5.19 (+40.7\%) \\
     & 128 & 12m02s & 6.06 (+64.2\%) \\
     & 64 & 11m03s & 6.60 (+78.9\%) \\
    \midrule
    \multirow{3}{*}{w/ VISA (Ours)} & 192 & 13m46s & 5.30 (+43.6\%) \\
     & 128 & 12m10s & 6.00 (+62.6\%) \\
     & 64 & 10m55s & 6.68 (+81.0\%) \\
    \toprule
    LLaVA-1.5-13B & 576 & 41m17s & 1.77 \\
    \midrule
    \multirow{3}{*}{w/ fastv} & 192 & 22m54s & 3.19 (+80.2\%) \\
     & 128 & 19m45s & 3.69 (+108.5\%) \\
     & 64 & 17m35s & 4.15 (+134.5\%) \\
    \midrule
    \multirow{3}{*}{w/ VISA (Ours)} & 192 & 22m49s & 3.20 (+80.8\%) \\
     & 128 & 19m46s & 3.69 (+108.5\%) \\
     & 64 & 17m28s & 4.18 (+136.2\%) \\

  \bottomrule
  \end{tabular}
  \label{table4}
\end{table}

\textbf{Model architectures.}
We validate the effectiveness of our method on various MLLM architectures, including LLaVA-1.5~\cite{ref6} with both 7B and 13B parameters, LLaVA-NeXT-7B~\cite{ref8} for high-resolution image inputs, and Video-LLaVA-7B~\cite{ref16} for video understanding. We maintain the same inference settings as the original code implementations for each model.

\begin{table*}[!t]
  \centering
  \setlength{\tabcolsep}{7pt}
  \renewcommand\arraystretch{1.2}
  \caption{Comparison with previous method on Video-LLaVA across 4 widely-used video understanding benchmarks.}
  \begin{tabular}{@{}lcccccccccccc@{}}
    \toprule
    &  &  & \multicolumn{2}{c}{TGIF-QA} & \multicolumn{2}{c}{MSVD-QA} & \multicolumn{2}{c}{MSRVTT-QA} & \multicolumn{2}{c}{ActivityNet-QA} & \multicolumn{2}{c}{Avg.} \\
    Method & Tokens & TFLOPs & Acc. & Score & Acc. & Score & Acc. & Score & Acc. & Score & Acc. & Score \\
    \midrule
    Video-LLaVA & 2048 & 30.63 & 19.1 & 2.50 & 70.8 & 3.96 & 57.4 & 3.50 & 43.8 & 3.33 & 100\% & 100\% \\
    \midrule
    FastV & \multirow{2}{*}{512} & \multirow{2}{*}{9.94} & 19.9 & 2.51 & 69.5 & 3.92 & 53.6 & 3.38 & 38.4 & 3.12 & 95.9\% & 97.4\% \\
    VISA (Ours) &  &  & 22.0 & 2.58 & 70.4 & 3.92 & 54.2 & 3.40 & 41.5 & 3.28 & \textbf{100.9\%} & \textbf{99.5\%} \\
    \midrule
    FastV & \multirow{2}{*}{256} & \multirow{2}{*}{6.71} & 15.7 & 2.44 & 67.3 & 3.86 & 52.9 & 3.36 & 36.6 & 3.03 & 88.2\% & \textbf{95.5\%} \\
    VISA (Ours) &  &  & 21.6 & 2.48 & 64.6 & 3.76 & 51.2 & 3.31 & 38.5 & 3.08 & \textbf{95.4\%} & 95.3\% \\
    \midrule
    FastV & \multirow{2}{*}{128} & \multirow{2}{*}{5.13} & 14.1 & 2.38 & 64.7 & 3.79 & 50.9 & 3.29 & 35.8 & 2.94 & 83.9\% & 93.3\% \\
    VISA (Ours) &  &  & 18.8 & 2.40 & 61.0 & 3.62 & 51.4 & 3.34 & 37.1 & 3.02 & \textbf{89.7\%} & \textbf{93.4\%} \\
    
  \bottomrule
  \end{tabular}
  \label{table5}
\end{table*}

\begin{table*}[!t]
  \centering
  \setlength{\tabcolsep}{3.5pt}
  \renewcommand\arraystretch{1.2}
  \caption{Ablation study of VTA, both methods adopt layer-wise compression strategy, with 192 visual tokens retained.}
  \begin{tabular}{@{}lccccccccccc@{}}
    \toprule
    Method & MM-Vet & MMB & MMB-CN & POPE & SQA-IMG & TextVQA & VQAv2 & VizWiz & MME & GQA & Avg. \\
    \midrule
    ToMe & 28.8 & 63.3 & 57.6 & 83.6 & 68.3 & 52.2 & 74.5 & 50.8 & 1446.7 & 58.5 & 96.6\% \\
    VTA (Ours) & 31.6 & 64.5 & 59.0 & 86.1 & 68.3 & 58.0 & 77.8 & 49.5 & 1499.1 & 60.5 & \textbf{100.0\%} \\
    
  \bottomrule
  \end{tabular}
  \label{table6}
\end{table*}

\begin{table*}[!t]
  \centering
  \setlength{\tabcolsep}{3.5pt}
  \renewcommand\arraystretch{1.2}
  \caption{Ablation study of using layer-wise compression strategy and GTS on VTA, with 128 visual tokens retained.}
  \begin{tabular}{@{}lccccccccccc@{}}
    \toprule
    Method & MM-Vet & MMB & MMB-CN & POPE & SQA-IMG & TextVQA & VQAv2 & VizWiz & MME & GQA & Avg. \\
    \midrule
    layer-wise & 23.0 & 64.0 & 58.2 & 84.8 & 68.2 & 56.0 & 75.6 & 45.7 & 1458.0 & 57.7 & 94.8\% \\
    GTS (Ours) & 30.2 & 64.8 & 58.0 & 82.4 & 67.8 & 57.3 & 76.4 & 49.2 & 1469.0 & 58.2 & \textbf{98.0\%} \\
    
  \bottomrule
  \end{tabular}
  \label{table7}
\end{table*}

\textbf{Comparison methods.}
We compare our method with FastV~\cite{ref11}, ToMe~\cite{ref12}, SparseVLM~\cite{ref36}, and PyramidDrop~\cite{ref46}.
FastV prunes visual tokens based on attention scores. ToMe introduces a straightforward visual information merging approach. SparseVLM initially prunes tokens based on the text prompt and subsequently suggests a token recycling mechanism based on feature averages to reduce information loss. PyramidDrop~\cite{ref46} divide the LLM decoding process into multiple stages, and drop part of the image tokens at the end of each stage based on the attention score of the visual tokens.
Since ToMe, SparseVLM, PyramidDrop, and our method compress visual tokens progressively, while FastV only prunes visual tokens in the second layer of LLM, in order to ensure a fair comparison, we adjust the corresponding hyperparameters of each method to ensure the same FLOPs as FastV which prunes visual tokens once in the second layer of LLM.

\subsection{Main Results}
We first evaluate our method and other competitors on LLaVA-1.5-7B across 10 multi-modal benchmarks. As depicted in Table\,\ref{table1}, our method attains the top overall performance across all compression ratio settings, while other methods experience significant performance degradation on certain datasets at higher compression rates due to the loss of visual information.

We also compare the results on the 13B version of LLaVA-1.5. As illustrated in Table\,\ref{table2}, our method consistently surpasses the previous approaches, which proves that our method also works well on models with larger parameters. Due to space limitations, we only present the average performance across 10 benchmarks.

\subsection{VISA with High Resolution}
By increasing the resolution of input images, some studies have achieved improved performance on many multimodal tasks~\cite{ref8, ref37, ref38}. However, high-resolution image inputs lead to longer visual token sequences, which further slows down the model's inference speed. We apply our method to LLaVA-NeXT-7B, which can handle up to 2880 visual tokens. Compared with other competitors, our method also excelled in high-resolution scenarios, which further proves the effectiveness of our method. The average performance on 10 benchmarks is shown in Table\,\ref{table3}. Our method maintains 91.1\% performance even with only 180 visual tokens retained.

\subsection{VISA with Video Understanding}
Video is another input modality converted into numerous visual tokens. In Video-LLaVA, 8 frames are extracted from a video input and these frames are finally converted into 2048 visual token inputs, showing a high degree of redundancy. Since ToMe cannot get normal model output when applied to Video-LLaVA, and The open source codes of SparseVLM and PyramidDrop lack implementation on Video-LLaVA, we only compare our method with FastV on Video-LLaVA. Due to the commercial API usage limits, we follow~\cite{ref11} to evaluate 1K samples from each benchmark in our experiments. As shown in Table\,\ref{table5}, our method can maintain 89.7\% model performance even with only 128 visual tokens retained. 

\begin{figure*}[!t]
  \centering
  \includegraphics[width=\textwidth]{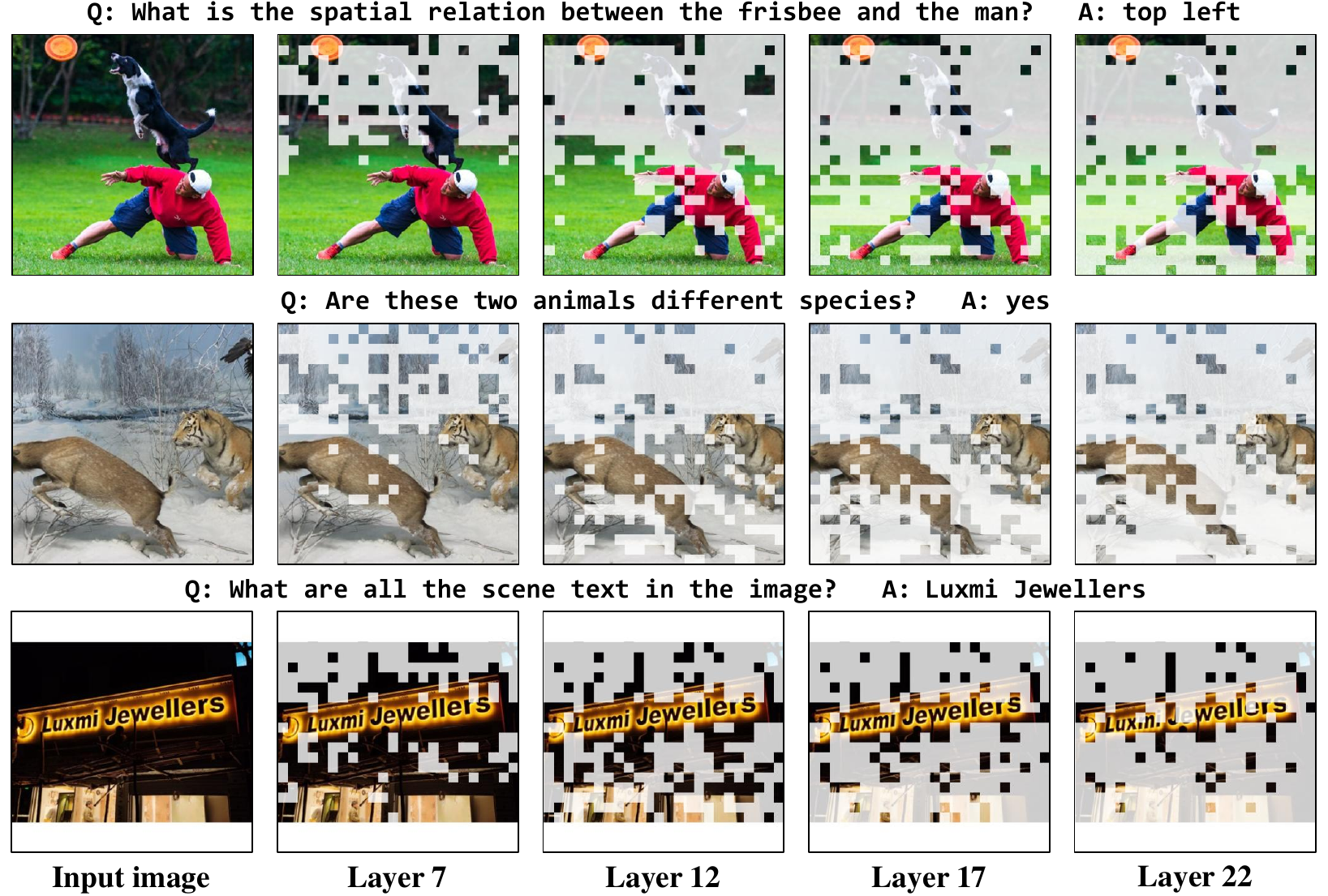}
  \caption{Visualization of the visual token aggregation process. Our method can preserve the crucial information related to the question prompt and progressively transform the image features into a compact representation.}
  \label{fig3}
\end{figure*}

\subsection{Inference Acceleration Analysis}
To showcase the efficiency of our method, we test the actual inference time of LLaVA-1.5 with VISA on the RTX 3090 GPU. The test benchmark is MMBench. Since MMBench necessitates the model to answer just one option per example, this eliminates the impact of the output sequence length on the inference time.
We also test the actual inference time of FastV at the same TFLOPs~(number of visual tokens retained) for comparison.  
As shown in Table\,\ref{table4}, our method can boost model throughput by 43.6\% on LLaVA-1.5-7B and by 80.8\% on LLaVA-1.5-13B with almost no loss in performance, significantly enhancing the model's inference speed.

\subsection{Ablation Study}

Our method primarily consists of visual token aggregation and group-wise token selection strategy. To showcase the efficacy of visual token aggregation, we contrast it with a basic token information merging method named ToMe. For a fair comparison, we employ the layer-wise compression strategy in both our method and ToMe. As depicted in Table\,\ref{table6}, our method preserves the original model performance even when retaining only 192 visual tokens. In contrast, ToMe experience a decline in performance due to the loss of visual information caused by simple feature averaging.
Meanwhile, we demonstrate the effectiveness of the group-wise token selection strategy in Table\,\ref{table7}. With only 128 visual tokens retained, GTS exceeds the layer-wise strategy by 3.2 percentage points.


\subsection{Visualization Results}
We visualize the pruned visual tokens at different layers in Figure\,\ref{fig3}. As the decoding process proceeds, unimportant visual tokens are progressively summarized into critical visual tokens, ultimately forming a compact representation of the input image. Additionally, our method can retain visual tokens which are related to the question prompt. For instance, in the first row of Figure\,\ref{fig3}, our method always retains visual tokens related to ``frisbee" and ``man". Meanwhile, in the third row of Figure\,\ref{fig3}, our method retains the key text area to ensure that the model can answer the question correctly.

\section{Conclusion}
We introduce a novel approach called group-wise visual token selection and aggregation~(VISA) to efficiently compress visual tokens in MLLMs. By effectively aggregating visual information through graph summarization, visual token aggregation~(VTA) significantly shortens visual token sequences in MLLMs, boosting MLLM inference speed with minimal impact on performance. Additionally, our proposed group-wise token selection strategy ensures the visual information extraction process more stable, especially under extreme compression setting, further improving VTA's performance.

\begin{acks}
This work was supported by the National Science Fund for Distinguished Young Scholars (No.62025603), the National Natural Science Foundation of China (No.U21B2037, No.U22B2051, No.U23A20383, No.U21A20472, No.62176222, No.62176223, No.62176226, No.62072386, No.62072387, No.62072389, No.62002305 and No.62272401), and the Natural Science Foundation of Fujian Province of China (No.2021J06003, No.2022J06001).
\end{acks}

\bibliographystyle{ACM-Reference-Format}
\bibliography{sample-base}

\onecolumn
\appendix

\section{Appendix}
In this appendix, we first compare our method with other competitors on image captioning task. Than we present ablation study on hyperparameter settings of our method. We also list the values of $p$ utilized for each compression rate settings. Finally, we present detailed comparison results on LLaVA-1.5-7B, LLaVA-1.5-13B and LLaVA-NeXT-7B across 10 multimodal benchmarks.

\subsection{Comparison on Image Captioning Task}
To assess our method's performance on image captioning task, we apply our method to LLaVA-1.5-7B and evaluate it on the nocaps benchmark. We report the CIDEr and ROUGE-L metrics. As depicted in Table\,\ref{table8}, our method outperforms previous approaches in seven of the eight comparisons., demonstrating its effectiveness in caption tasks.

\begin{table*}[h]
  \centering
  \setlength{\tabcolsep}{7pt}
  \renewcommand\arraystretch{1.2}
  \caption{Comparison with previous methods on the nocaps benchmark, the CIDEr and ROUGE-L metrics are reported.}
  \begin{tabular}{@{}lcccccccccccccc@{}}
    \toprule
    & \multicolumn{2}{c}{Tokens: 288} & \multicolumn{2}{c}{Tokens: 192} & \multicolumn{2}{c}{Tokens: 128} & \multicolumn{2}{c}{Tokens: 64} \\
    Method & CIDEr & ROUGE-L & CIDEr & ROUGE-L & CIDEr & ROUGE-L & CIDEr & ROUGE-L \\
    \midrule
    FastV & 96.3 & 58.1 & 94.9 & 57.6 & \textbf{91.1} & 56.4 & 75.2 & 51.8 \\
    ToMe & 94.7 & 57.6 & 90.5 & 56.6 & 82.3 & 54.1 & 67.6 & 50.7 \\
    SparseVLM & 96.3 & 57.7 & 94.6 & 57.5 & 88.9 & 55.6 & 70.6 & 51.1 \\
    PyramidDrop & 96.5 & 57.4 & 94.1 & 57.3 & 89.7 & 56.2 & 74.3 & 51.6 \\
    VISA (Ours) & \textbf{98.3} & \textbf{58.4} & \textbf{95.4} & \textbf{57.7} & 90.6 & \textbf{56.6} & \textbf{76.0} & \textbf{53.8} \\
  \bottomrule
  \end{tabular}
  \label{table8}
\end{table*}

\subsection{Ablation Study of Hyperparameter Settings.}
Our method mainly consists of three hyperparameters: $\alpha$ in visual tokens aggregation, the number of layers $S$ contained in each group, and the last $M$ layers of each group employed to average attention score. We conduct ablation experiments on LLaVA-1.5-7B with 128 visual tokens kept to determine the best hyperparameter settings. 
As shown in Table\,\ref{table9}, with $\alpha=0.1$, $S=5$, and $M=2$, our method achieve the best average model performance.

\begin{table*}[h]
  \centering
  \setlength{\tabcolsep}{6pt}
  \renewcommand\arraystretch{1.2}
  \caption{Ablation study of hyperparameter settings, when one hyperparameter is changed, the other hyperparameters remain unchanged.}
  \begin{tabular}{@{}lccccccccccc@{}}
    \toprule
    Method & MM-Vet & MMB & MMB-CN & POPE & SQA-IMG & TextVQA & VQAv2 & VizWiz & MME & GQA & Avg. \\
    \midrule
    $\alpha=0.1$ & 30.2 & 64.8 & 58.0 & 82.4 & 67.8 & 57.3 & 76.4 & 49.2 & 1469.0 & 58.2 & \textbf{98.0\%} \\
    $\alpha=0.2$ & 27.8 & 64.5 & 58.2 & 82.1 & 67.9 & 57.0 & 76.1 & 49.5 & 1459.5 & 58.5 & 97.2\% \\
    $\alpha=0.3$ & 29.3 & 63.7 & 57.6 & 82.7 & 68.3 & 56.6 & 76.3 & 49.9 & 1452.1 & 58.4 & 97.5\% \\

    \midrule
    $S=4$ & 28.9 & 63.7 & 57.6 & 82.3 & 68.1 & 57.0 & 76.0 & 49.4 & 1443.6 & 58.4 & 97.2\% \\
    $S=5$ & 30.2 & 64.8 & 58.0 & 82.4 & 67.8 & 57.3 & 76.4 & 49.2 & 1469.0 & 58.2 & \textbf{98.0\%} \\
    $S=6$ & 27.8 & 64.0 & 58.5 & 81.9 & 67.8 & 57.2 & 76.6 & 49.6 & 1478.2 & 57.9 & 97.2\% \\

    \midrule
    $M=1$ & 28.6 & 64.2 & 58.4 & 82.5 & 67.5 & 56.9 & 75.8 & 48.1 & 1465.7 & 58.5 & 97.1\% \\
    $M=2$ & 30.2 & 64.8 & 58.0 & 82.4 & 67.8 & 57.3 & 76.4 & 49.2 & 1469.0 & 58.2 & \textbf{98.0\%} \\
    $M=3$ & 28.9 & 64.3 & 58.4 & 82.3 & 67.6 & 57.4 & 76.2 & 48.9 & 1479.6 & 58.2 & 97.5\% \\
  \bottomrule
  \end{tabular}
  \label{table9}
\end{table*}

\subsection{Details of the Number of Retained Visual Tokens in the LLM Decoding Process}
Our approach involves progressively shortening the length of visual token sequence. Specifically, for an input visual token sequence of length $n$, given the proportion $p$ of visual tokens to be retained for each visual token aggregation module, the number of visual tokens after the $k$-th group is $n \cdot p^k$. Taking LLaVA-1.5-13B as an example, We list the values of $p$ utilized at each compression rate settings in the main experiment in Table\,\ref{table10}.

\begin{table*}[h]
  \centering
  \setlength{\tabcolsep}{10pt}
  \renewcommand\arraystretch{1.25}
  \caption{The values of $p$ utilized at each compression rate settings in the main experiment for LLaVA-1.5-13B, the number of layers contained in each group is 7. Each column in the table represents the TFLOPs of our progressive compression strategy aligned with FastV's one-time pruning strategy at the second layer.}
  \begin{tabular}{@{}ccccc@{}}
    \toprule
    Tokens & 288 & 192 & 128 & 64 \\
    \midrule
    TFLOPs & 11.17 & 8.82 & 7.26 & 5.70 \\
    \midrule
    $p$ & 0.790 & 0.675 & 0.560 & 0.380 \\
  \bottomrule
  \end{tabular}
  \label{table10}
\end{table*}

\subsection{Detailed Experimental Results}
We provide the complete comparison results on LLaVA-1.5-7B in Table\,\ref{table11}.
Meanwhile, we provide detailed comparison results on LLaVA-1.5-13B and LLaVA-NeXT-7B in Table\,\ref{table12} and Table\,\ref{table13}, respectively. Our method consistently outperform previous methods in all kept visual token number settings. Notably, with only 180 visual tokens retained, our method can still maintain 91.1\% performance on LLaVA-NeXT-7B. 
The top overall performance validate the effectiveness of our method on larger models and high resolution scenes.

\begin{table*}[h]
  \centering
  \setlength{\tabcolsep}{3.5pt}
  \renewcommand\arraystretch{1.25}
  \caption{Comparison with previous methods on LLaVA-1.5-7B across 10 multimodal benchmarks.}
  \begin{tabular}{@{}lcccccccccccc@{}}
    \toprule
    Method & TFLOPs & MM-Vet & MMB & MMB-CN & POPE & SQA-IMG & TextVQA & VQAv2 & VizWiz & MME & GQA & Avg. \\
    \midrule
    LLaVA-1.5-7B & 9.43 & 31.1 & 64.3 & 58.3 & 85.9 & 66.8 & 58.2 & 78.5 & 50.0 & 1510.7 & 62.0 & 100\% \\
    \midrule
    \rowcolor{gray!20}\multicolumn{13}{c}{\emph{Retain 288 Tokens}} \\
    FastV & \multirow{5}{*}{5.79} & 29.7 & 64.0 & 58.4 & 82.3 & 67.6 & 58.1 & 77.7 & 50.5 & 1514.9 & 60.1 & 98.9\% \\
    ToMe &  & 30.2 & 63.7 & 58.1 & 85.0 & 67.7 & 54.6 & 76.7 & 51.2 & 1452.6 & 59.9 & 98.3\% \\
    SpareseVLM &  & 31.5 & 63.1 & 56.9 & 83.9 & 67.6 & 57.5 & 76.7 & 51.0 & 1458.8 & 58.8 & 98.6\% \\
    PyramidDrop &  & 30.8 & 65.0 & 58.3 & 85.8 & 68.3 & 57.5 & 78.0 & 50.5 & 1513.0 & 61.1 & 100.0\% \\
    VISA (Ours) &  & 33.1 & 64.8 & 59.0 & 85.7 & 67.8 & 58.0 & 78.2 & 49.8 & 1506.2 & 61.4 & \textbf{100.7\%} \\
    \midrule
    \rowcolor{gray!20}\multicolumn{13}{c}{\emph{Retain 192 Tokens}} \\
    FastV & \multirow{5}{*}{4.59} & 29.8 & 64.2 & 57.6 & 77.7 & 67.7 & 57.9 & 75.9 & 50.8 & 1485.1 & 57.9 & 97.6\% \\
    ToMe &  & 28.8 & 63.3 & 57.6 & 83.6 & 68.3 & 52.2 & 74.5 & 50.8 & 1446.7 & 58.5 & 96.6\% \\
    SpareseVLM &  & 31.2 & 62.5 & 56.1 & 83.6 & 69.1 & 56.1 & 75.6 & 51.1 & 1406.2 & 57.6 & 97.5\% \\
    PyramidDrop &  & 31.2 & 63.3 & 56.8 & 82.3 & 68.8 & 56.5 & 75.1 & 51.1 & 1479.7 & 57.3 & 98.0\% \\
    VISA (Ours) &  & 31.7 & 64.5 & 58.7 & 85.7 & 67.9 & 57.4 & 77.6 & 49.7 & 1503.3 & 60.3 & \textbf{99.8\%} \\
    \midrule
    \rowcolor{gray!20}\multicolumn{13}{c}{\emph{Retain 128 Tokens}} \\
    FastV & \multirow{5}{*}{3.80} & 26.8 & 63.3 & 56.3 & 72.0 & 68.0 & 56.9 & 73.2 & 51.3 & 1442.9 & 55.8 & 94.6\% \\
    ToMe &  & 25.3 & 61.8 & 56.5 & 81.4 & 67.8 & 50.8 & 71.8 & 50.3 & 1378.8 & 57.8 & 93.4\% \\
    SpareseVLM &  & 30.1 & 60.0 & 53.9 & 80.5 & 67.1 & 54.9 & 73.8 & 52.2 & 1386.9 & 56.0 & 95.2\% \\
    PyramidDrop &  & 29.4 & 61.6 & 56.6 & 82.3 & 68.4 & 56.6 & 72.9 & 51.0 & 1417.3 & 57.1 & 96.4\% \\
    VISA (Ours) &  & 30.2 & 64.8 & 58.0 & 82.4 & 67.8 & 57.3 & 76.4 & 49.2 & 1469.0 & 58.2 & \textbf{98.0\%} \\
    \midrule
    \rowcolor{gray!20}\multicolumn{13}{c}{\emph{Retain 64 Tokens}} \\
    FastV & \multirow{5}{*}{3.01} & 27.3 & 59.5 & 52.1 & 59.2 & 68.1 & 55.2 & 66.3 & 51.8 & 1237.1 & 51.6 & 88.9\% \\
    ToMe &  & 19.9 & 48.9 & 37.9 & 50.2 & 68.8 & 45.4 & 58.8 & 49.1 & 1059.0 & 49.8 & 76.8\% \\
    SpareseVLM &  & 22.3 & 56.2 & 46.5 & 75.1 & 62.2 & 51.8 & 68.2 & 51.3 & 1201.4 & 52.7 & 86.2\% \\
    PyramidDrop &  & 25.9 & 58.8 & 50.5 & 55.9 & 69.0 & 50.6 & 69.2 & 50.7 & 1309.2 & 47.5 & 86.9\% \\
    VISA (Ours) &  & 25.0 & 62.1 & 57.3 & 77.6 & 67.9 & 55.6 & 74.1 & 47.9 & 1420.6 & 56.2 & \textbf{93.8\%} \\
  \bottomrule
  \end{tabular}
  \label{table11}
\end{table*}

\begin{table*}[h]
  \centering
  \setlength{\tabcolsep}{3.5pt}
  \renewcommand\arraystretch{1.25}
  \caption{Comparison with previous methods on LLaVA-1.5-13B across 10 multimodal benchmarks.}
  \begin{tabular}{@{}lcccccccccccc@{}}
    \toprule
    Method & TFLOPs & MM-Vet & MMB & MMB-CN & POPE & SQA-IMG & TextVQA & VQAv2 & VizWiz & MME & GQA & Avg. \\
    \midrule
    LLaVA-1.5-13B & 18.31 & 36.1 & 67.7 & 63.6 & 85.9 & 71.6 & 61.3 & 80.0 & 53.6 & 1531.3 & 63.3 & 100\% \\
    \midrule
    \rowcolor{gray!20}\multicolumn{13}{c}{\emph{Retain 288 Tokens}} \\
    FastV & \multirow{5}{*}{11.17} & 34.7 & 67.7 & 61.8 & 85.2 & 71.4 & 60.8 & 79.5 & 54.2 & 1543.1 & 62.6 & 99.2\% \\
    ToMe &  & 35.2 & 67.0 & 61.8 & 86.0 & 70.8 & 57.4 & 78.4 & 53.8 & 1529.5 & 61.1 & 98.1\% \\
    SpareseVLM &  & 36.5 & 66.7 & 61.9 & 84.6 & 71.0 & 59.5 & 78.5 & 53.1 & 1497.4 & 59.9 & 98.1\% \\
    PyramidDrop &  & 36.7 & 67.8 & 62.1 & 86.3 & 71.6 & 60.5 & 79.5 & 52.9 & 1493.5 & 61.4 & 99.1\% \\
    VISA (Ours) &  & 36.4 & 67.8 & 62.0 & 87.2 & 71.2 & 61.1 & 79.4 & 53.6 & 1515.1 & 62.3 & \textbf{99.6\%} \\
    \midrule
    \rowcolor{gray!20}\multicolumn{13}{c}{\emph{Retain 192 Tokens}} \\
    FastV & \multirow{5}{*}{8.82} & 34.2 & 67.5 & 61.3 & 82.7 & 71.7 & 60.5 & 78.4 & 54.4 & 1489.8 & 61.0 & 97.9\% \\
    ToMe &  & 33.1 & 66.2 & 62.0 & 85.5 & 71.1 & 56.1 & 77.4 & 54.5 & 1528.6 & 60.2 & 97.1\% \\
    SpareseVLM &  & 35.9 & 66.3 & 61.3 & 82.9 & 71.1 & 59.1 & 77.1 & 53.2 & 1466.0 & 58.7 & 97.0\% \\
    PyramidDrop &  & 36.8 & 67.8 & 61.8 & 85.6 & 72.5 & 59.8 & 78.8 & 52.3 & 1516.8 & 61.0 & 98.9\% \\
    VISA (Ours) &  & 36.7 & 67.8 & 61.4 & 86.8 & 71.2 & 60.5 & 79.0 & 53.4 & 1511.9 & 61.9 & \textbf{99.2\%} \\
    \midrule
    \rowcolor{gray!20}\multicolumn{13}{c}{\emph{Retain 128 Tokens}} \\
    FastV & \multirow{5}{*}{7.26} & 31.4 & 66.5 & 61.3 & 77.6 & 72.5 & 59.7 & 76.5 & 55.0 & 1487.4 & 59.3 & 96.0\% \\
    ToMe &  & 24.8 & 66.1 & 60.8 & 82.5 & 71.2 & 53.5 & 72.7 & 53.0 & 1514.4 & 57.6 & 92.4\% \\
    SpareseVLM &  & 34.0 & 64.3 & 60.2 & 81.1 & 71.4 & 57.9 & 75.2 & 53.1 & 1467.5 & 57.9 & 95.3\% \\
    PyramidDrop &  & 35.0 & 67.6 & 60.8 & 84.7 & 71.2 & 59.7 & 78.1 & 52.7 & 1461.9 & 59.4 & 97.3\% \\
    VISA (Ours) &  & 34.7 & 67.8 & 61.6 & 85.1 & 72.1 & 60.2 & 78.4 & 52.9 & 1506.0 & 60.3 & \textbf{98.1\%} \\
    \midrule
    \rowcolor{gray!20}\multicolumn{13}{c}{\emph{Retain 64 Tokens}} \\
    FastV & \multirow{5}{*}{5.70} & 31.8 & 63.1 & 57.1 & 69.0 & 72.9 & 56.1 & 71.1 & 55.2 & 1364.4 & 55.4 & 91.3\% \\
    ToMe &  & 26.1 & 59.5 & 51.4 & 71.5 & 70.7 & 50.1 & 67.1 & 54.5 & 1287.0 & 55.3 & 86.2\% \\
    SpareseVLM &  & 30.1 & 62.5 & 55.5 & 76.1 & 71.9 & 54.9 & 69.4 & 52.9 & 1360.5 & 54.8 & 90.2\% \\
    PyramidDrop &  & 28.7 & 65.9 & 59.5 & 79.0 & 72.6 & 57.0 & 74.9 & 51.6 & 1419.8 & 56.8 & 92.9\% \\
    VISA (Ours) &  & 29.1 & 66.0 & 60.3 & 79.5 & 73.9 & 59.2 & 75.5 & 52.0 & 1468.3 & 57.4 & \textbf{94.3\%} \\
  \bottomrule
  \end{tabular}
  \label{table12}
\end{table*}

\begin{table*}[h]
  \centering
  \setlength{\tabcolsep}{3.5pt}
  \renewcommand\arraystretch{1.25}
  \caption{Comparison with previous methods on LLaVA-NeXT-7B across 10 multimodal benchmarks.}
  \begin{tabular}{@{}lcccccccccccc@{}}
    \toprule
    Method & TFLOPs & MM-Vet & MMB & MMB-CN & POPE & SQA-IMG & TextVQA & VQAv2 & VizWiz & MME & GQA & Avg. \\
    \midrule
    LLaVA-NeXT-7B & 43.61 & 43.9 & 67.4 & 60.6 & 86.5 & 70.1 & 64.9 & 81.8 & 57.6 & 1519.0 & 64.2 & 100\% \\
    \midrule
    \rowcolor{gray!20}\multicolumn{13}{c}{\emph{Retain 720 Tokens}} \\
    FastV & \multirow{5}{*}{13.41} & 42.8 & 67.4 & 59.9 & 84.5 & 68.9 & 60.3 & 80.6 & 56.7 & 1501.5 & 63.3 & 98.0\% \\
    ToMe &  & 32.4 & 64.9 & 55.6 & 85.3 & 67.2 & 50.7 & 77.9 & 54.6 & 1407.8 & 60.6 & 91.2\% \\
    SpareseVLM &  & 38.0 & 63.8 & 57.0 & 85.7 & 67.5 & 58.1 & 78.9 & 55.6 & 1446.1 & 60.9 & 94.3\% \\
    PyramidDrop &  & 40.0 & 67.4 & 60.0 & 86.4 & 68.4 & 59.7 & 79.9 & 56.6 & 1491.2 & 62.5 & 97.1\% \\
    VISA (Ours) &  & 45.5 & 68.3 & 60.7 & 87.2 & 69.0 & 61.0 & 80.9 & 57.2 & 1508.5 & 63.5 & \textbf{99.5\%} \\
    \midrule
    \rowcolor{gray!20}\multicolumn{13}{c}{\emph{Retain 360 Tokens}} \\
    FastV & \multirow{5}{*}{8.83} & 38.9 & 66.0 & 57.3 & 79.7 & 68.4 & 59.0 & 77.6 & 55.0 & 1413.4 & 60.6 & 94.0\% \\
    ToMe &  & 30.0 & 62.1 & 54.6 & 82.4 & 67.7 & 47.1 & 74.4 & 52.2 & 1343.2 & 59.4 & 87.8\% \\
    SpareseVLM &  & 35.4 & 63.4 & 54.8 & 83.4 & 67.8 & 55.5 & 75.3 & 53.5 & 1401.1 & 58.2 & 91.2\% \\
    PyramidDrop &  & 36.2 & 65.1 & 57.5 & 81.8 & 69.2 & 56.9 & 77.3 & 55.7 & 1449.7 & 58.8 & 93.3\% \\
    VISA (Ours) &  & 39.0 & 66.4 & 58.7 & 86.3 & 69.2 & 59.1 & 79.2 & 55.3 & 1439.0 & 61.0 & \textbf{95.6\%} \\
    \midrule
    \rowcolor{gray!20}\multicolumn{13}{c}{\emph{Retain 180 Tokens}} \\
    FastV & \multirow{5}{*}{6.58} & 30.0 & 61.4 & 53.9 & 67.0 & 68.5 & 55.6 & 69.8 & 52.3 & 1223.3 & 54.7 & 85.1\% \\
    ToMe &  & 24.6 & 40.1 & 34.0 & 73.6 & 64.1 & 34.8 & 52.4 & 37.3 & 932.9 & 44.4 & 66.1\% \\
    SpareseVLM &  & 22.6 & 55.2 & 48.1 & 69.2 & 66.7 & 45.2 & 66.7 & 48.1 & 1155.5 & 50.1 & 77.7\% \\
    PyramidDrop &  & 29.3 & 64.2 & 56.3 & 77.2 & 67.4 & 55.0 & 73.4 & 54.1 & 1375.9 & 57.2 & 88.8\% \\
    VISA (Ours) &  & 33.1 & 65.0 & 55.7 & 83.1 & 68.5 & 56.5 & 76.0 & 53.2 & 1367.3 & 58.4 & \textbf{91.1\%} \\
  \bottomrule
  \end{tabular}
  \label{table13}
\end{table*}

\end{document}